\documentclass[conference]{IEEEtran}   

\usepackage{amsmath, amssymb, amsthm, bm}
\usepackage{graphicx}
\usepackage{subcaption}
\usepackage{booktabs}
\usepackage{hyperref}
\usepackage{algorithm}
\usepackage[noend]{algpseudocode}
\usepackage{multirow}
\usepackage{lipsum}

\begin{document}
\title{Causal and Federated Multimodal Learning for Cardiovascular Risk Prediction under Heterogeneous Populations}

\author{
  Rohit Kaushik\textsuperscript{1,*} \hspace{1em} 
  Eva Kaushik\textsuperscript{2} \\[4pt]
  \textsuperscript{1}\,Data Analyst, Hanson Professional Services, USA\\
  \textsuperscript{2}\,Doctorate, Data Science, University of Tennessee, Knoxville (GRA, Oak Ridge National Laboratory, USA)\\[4pt]
  *\textit{Corresponding author:} \texttt{kaushikrohit2024@gmail.com}
}

\maketitle

\begin{abstract}

Cardiovascular disease (CVD) continues to be the major cause of death globally, calling for predictive models that not only handle diverse and high-dimensional biomedical signals but also maintain interpretability and privacy. We create a single multimodal learning framework that integrates cross, modal transformers with graph neural networks and causal representation learning to measure personalized CVD risk. The model combines genomic variation, cardiac MRI, ECG waveforms, wearable streams, and structured EHR data to predict risk while also implementing causal invariance constraints across different clinical subpopulations.

To maintain transparency, we employ SHAP based feature attribution, counterfactual explanations and causal latent alignment for understandable risk factors. Besides, we position the design in a federated, privacy, preserving optimization protocol and establish rules for convergence, calibration and uncertainty quantification under distributional shift.

Experimental studies based on large-scale biobank and multi institutional datasets reveal state discrimination and robustness, exhibiting fair performance across demographic strata and clinically distinct cohorts. This study paves the way for a principled approach to clinically trustworthy, interpretable and privacy respecting CVD prediction at the population level.
\end{abstract}


\section{Introduction}
Cardiovascular disease (CVD) is still the main reason for early death globally and is responsible for over 18 million deaths each year \cite{ref1}. To prevent CVD effectively, personal risk prediction must come first. This would help clinical decision, making be more effective and healthcare resources be used in a better way. One limitation of classical risk calculators that characterize the state of the art, e.g., the Framingham Risk Score  \cite{ref2}, is the dependence on a limited set of demographic and clinical variables. They also show that the same disease condition biases (sensitivity/population) and that the results are not compatible outside the cohorts on which the calculators were trained. Biomedical data taking has been rapidly progressing, and now it is possible to realize large, scale genomic sequencing, cardiac magnetic resonance imaging (MRI), electrocardiogram (ECG) waveforms, continuous wearable sensing, and longitudinal electronic health records (EHR) for healthy human characterization. These multimodal data streams provide a wealth of information to build personalized CVD risk models that are comprehensive. Nevertheless, some critical issues remain unsolved, (1) principled integration of heterogeneous modalities with different temporal and structural characteristics, (2) learning representations that transfer to new institutions and geographic populations, (3) holding patient data confidential and in agreement with privacy laws when doing distributed training, and (4) interpretability, fairness, and clinical relevance of downstream predictions. A recent effort in federated learning  \cite{ref3} and causal representation learning  \cite{ref4} has touched upon issues partly therein. However, these represent sizable components of the problem and are not complete solutions, usually in isolation and partly there.

\subsection{Objectives} The primary goal of this research is to construct a single justifiable framework for predicting cardiovascular risk, which: \begin{enumerate} \item \textbf{Multimodal integration:} Harmonizes the use of genomic profiles, cardiac imaging, ECG waveforms, longitudinal EHR and wearable, sensor data through cross, modal transformer architectures and graph based representations. \item \textbf{Privacy, preserving learning:} Facilitates collaborative training across different institutions through federated and split learning, at the same time, observing regulatory requirements and lessening the chances of information leakage. \item \textbf{Causal and clinically interpretable inference:} Generates explanations that are causally aligned, fair across demographic subgroups and have clinically interpretable attribution mechanisms. \item \textbf{Theoretical rigor:} Ensures formal provisions regarding optimization convergence uncertainty quantification, and invariance of learned representations due to changes in data, generating mechanisms. \end{enumerate}

\section{Literature Review} 
\subsection{Machine Learning for Cardiovascular Prediction} Initial attempts of machine learning on cardiovascular disease (CVD) prediction mainly used structured features from the electronic health records (EHR) and classical models like logistic regression, support vector machines, and random forests \cite{ref5}. While these methods were successful in limited environments, they still mostly disregarded high, resolution bio-signals and imaging data that can provide deeper insight into the pathophysiologic process. Thanks to the deep learning revolution, convolutional and recurrent neural networks have been able to learn directly from ECG waveforms and cardiac imaging with impressive results \cite{ref6}. Multimodal fusion is a recent attempt at combining disparate data streams in order to enhance predictive power. Nevertheless, these techniques frequently have difficulties in representation alignment, handling missing modalities as well as giving clinically meaningful interpretations. 
\subsection{Federated and Privacy, Preserving Learning} Federated learning enables the training of a collaborative model across different institutions, while the data remains local and the need for sharing raw patient information is avoided. In addition to split learning, secure aggregation, and differential privacy that are complementary techniques, they provide more protection against membership inference and data leakage \cite{ref7}. However, there are still very few fully realized federated frameworks that can handle heterogeneous multimodal inputs, asynchronous participation and clinical deployment. 

\subsection{Causal Inference and Explainable AI} For clinical deployment, the model needs to be not only accurate but also transparent and robust against confounding. Post, hoc interpretability tools such as SHAP and LIME can attribute

\subsection{Graph Neural Networks and Population, Level Modeling} In addition to patient level feature extraction and studies have recently employed graph neural networks (GNNs) to identify the relational structure across cohorts. Typically, graphs in clinical scenarios depict patient similarity, shared comorbidities, or care pathways and thus, allow message passing among the related subpopulations. GNN, based models have shown power in phenotype discovery, prediction of adverse events, and the final goal of merging heterogeneous biomedical knowledge graphs \cite{ref8}. Nevertheless, most of the approaches presented are based on assumptions of static graphs. They rely on manual graph construction and hardly ever consider factors such as fairness, causal validity, or privacy during distributed training. Moreover, the issue of incorporating dynamically evolving graphs that resemble temporal interactions and institutional heterogeneity is still an unsettled question in cardiovascular risk modeling. 

\subsection{Robustness, Fairness, and Out of Distribution Generalization} Clinical models that find their way into practical use have to remain reliable across demographic subgroups, clinical sites, and with the changes in practice patterns over time. Investigations have already been done, which have demonstrated that there are inherent biases in risk scores and machine, learning systems that cause different performances to be distributed unevenly across race, sex, and socioeconomic strata \cite{ref9}. Nevertheless, these methods have also been shown to lead to a trade off between the fairness achieved and the accuracy or interpretability of the system. On the other hand, uncertainty estimation methods that comprise Bayesian neural networks and deep ensembles create possibilities for decision support that is in line with real, life situations, but coordination of such methods particularly in federated and multimodal environments remains a challenge and they are still largely ignored  \cite{ref10}.

\section{Materials and Methods}

\subsection{Datasets} The framework is built to use rich, high-dimensional and multi-modal data sources that collectively capture human cardiovascular health on a large, scale basis: \begin{itemize} \item \textbf{UK Biobank:} More than 50, 000 subjects with cardiac MRI, ECG recordings, detailed EHR, and genome wide SNP profiles. This dataset makes the cross modal mapping and phenotyping at the population level possible. \item \textbf{MIT, BIH Arrhythmia and PhysioNet Repositories:} Extremely detailed ECG and wearable sensor data streams along with expert annotated arrhythmia events for temporal pattern learning and signal level feature extraction. \item \textbf{Federated Multi Hospital EHR:}  \cite{ref11} identified medical records from hospitals on four continents, harmonized for interoperability and processed under tight federated learning and differential privacy protocols, thus allowing distributed model training without exposing raw data. \item \textbf{Synthetic Augmentation Datasets:} Supplementary sets of simulated ECG and MRI used for testing the model's robustness when facing out of distribution data and fairness under rare phenotypes scenarios. \end{itemize}

\subsection{Preprocessing}
\textbf{Imaging:} All MRI volumes were resampled to a uniform voxel grid and normalized. Self-supervised ViT/ResNet modules were trained on segmentation masks and reconstruction tasks to produce embeddings $\mathbf{X}_{MRI} \in \mathbb{R}^{d_{MRI}}$.  

\textbf{ECG Signals:} Raw time-series $\mathbf{X}_{ECG}(t)$ were denoised via variational autoencoders, segmented into individual heartbeats, and embedded via CNN-LSTM architectures:  
\[
\mathbf{h}_{ECG}^{(i)} = \text{LSTM}(\text{CNN}(\mathbf{x}_{ECG}^{(i)})) \in \mathbb{R}^{d_{ECG}}
\]  

\textbf{Genomics:} SNPs were encoded as $\mathbf{X}_{G} \in \{0,1,2\}^{p}$, where $p$ is the number of loci. Polygenic risk scores (PRS) and rare variant aggregation were computed using standard additive models:  
\[
PRS_i = \sum_{j=1}^{p} \beta_j \cdot X_{G_{ij}}
\]

\textbf{Structured EHR:} Clinical, demographic, and lifestyle variables were embedded as $\mathbf{X}_{EHR} \in \mathbb{R}^{d_{EHR}}$ via learned dense representations.

\subsection{Model Architecture}

\subsubsection{Cross-Modal Transformer (XMT)}
The XMT fuses heterogeneous embeddings into a unified latent space $\mathbf{Z}$:
\[
\mathbf{Z} = \text{XMT}([\mathbf{X}_{ECG}, \mathbf{X}_{MRI}, \mathbf{X}_{EHR}, \mathbf{X}_{G}])
\]  
with multi-head self-attention:
\[
\text{Attention}(\mathbf{Q},\mathbf{K},\mathbf{V}) = \text{softmax}\left(\frac{\mathbf{Q}\mathbf{K}^\top}{\sqrt{d_k}}\right) \mathbf{V}
\]

Each modality embedding is linearly projected to query, key, and value spaces. Cross-modal attention allows the network to learn interactions between ECG morphology, imaging biomarkers, and genetic profiles.

\subsubsection{Graph Attention Network (GAT)}
To capture rare phenotypes, we construct a patient similarity graph $\mathcal{G}=(\mathcal{V}, \mathcal{E})$, with nodes representing patients and edges weighted by demographic/genomic similarity. Node features $\mathbf{h}_i$ are updated via attention:

\subsubsection{Federated Learning Optimization}
Let $K$ hospitals hold local datasets $\mathcal{D}_k$ of size $n_k$. Local updates:
\[
\theta_k^{t+1} = \theta_k^t - \eta \nabla_\theta \mathcal{L}(\theta_k; \mathcal{D}_k)
\]  
Global aggregation via FedAvg:
\[
\theta^{t+1} = \sum_{k=1}^{K} \frac{n_k}{N} \theta_k^{t+1}, \quad N = \sum_{k=1}^K n_k
\]

\subsubsection{Causal Latent Alignment}
We enforce invariance of embeddings across confounding variables $C$ using an instrumental variable regularization:
\[
\mathcal{L}_{causal} = \sum_{i,j} \| I(Z_i; Z_j \mid C) - \kappa \|_2^2
\]  
where $I( Z\mid C)$ is conditional mutual information and $\kappa$ is a target invariant value. This ensures model predictions are robust to distributional shifts.

\subsection{Combined Loss Function}
The total objective combines prediction, cross-modal mutual information, causal alignment, and regularization:
\[
\mathcal{L}_{total} = \mathbb{E}[\ell(y, \hat{y})] - \lambda \sum_{i,j} I(Z_i; Z_j) + \mu \mathcal{L}_{causal} + \gamma \|\theta\|_2^2
\]

\subsection{Training Algorithm}
\begin{algorithm}[H]
\caption{Federated Multimodal Training with Causal Alignment}
\begin{algorithmic}[1]
\For{each round $t=1:T$}
    \For{each hospital $k$ in parallel}
        \State Compute modality embeddings $\mathbf{X}_{ECG}, \mathbf{X}_{MRI}, \mathbf{X}_{EHR}, \mathbf{X}_{G}$
        \State Fuse embeddings via XMT to $\mathbf{Z}$
        \State Update $\theta_k$ using $\mathcal{L}_{total}$
    \EndFor
    \State Aggregate $\theta^{t+1} \gets \sum_{k} \frac{n_k}{N} \theta_k^{t+1}$
\EndFor
\end{algorithmic}
\end{algorithm}

\subsection{Statistical Validation and Fairness}
Model performance and robustness are assessed using detailed statistical and fairness metrics:
\begin{itemize}
    \item \textbf{Cross-validation:} Stratified 10, fold CV and leave, site, out validation evaluate generalization both across samples and across \cite{ref12} 
    \item \textbf{Calibration:} Calibration curves, Brier scores, and expected calibration error (ECE) measure the closeness between predicted and observed risks.  
    \item \textbf{Fairness:}  Demographic parity, equality of opportunity, and disaggregated AUC/F1 for sex, age, and ethnicity ensure fair predictive performance. 
    \item \textbf{Uncertainty:} Performance on out of distribution cohorts (e.g., unseen hospitals or rare phenotypes) gauges causal invariance and predictive stability.This formalism guarantees that predictions are not only precise but also interpretable, fair and trustworthy for clinical use.
\end{itemize}

\section{Mathematical Framework}

\subsection{Cross-Modal Representation Learning}
Let $\mathbf{X} = \{X_m\}_{m=1}^M$ be embeddings from $M$ modalities. Cross-modal attention learns latent vectors $\mathbf{Z}$ that maximize predictive mutual information:
\[
\max_\theta \sum_{i=1}^N I(\mathbf{Z}_i; Y_i) - \lambda \sum_{m<n} I(Z_i^m; Z_i^n)
\]

\subsection{Graph Neural Message Passing}
For patient graph $\mathcal{G}$, node updates:
\[
\begin{aligned}
\mathbf{h}_i^{l+1} &= 
\sigma\!\left(
  \sum_{j\in\mathcal{N}(i)} \alpha_{ij}^l W^l \mathbf{h}_j^l + b^l
\right), \\
\alpha_{ij}^l &=
\frac{\exp\!\big(\mathrm{LeakyReLU}(a^\top [W\mathbf{h}_i \,\Vert\, W\mathbf{h}_j])\big)}
     {\sum_{k \in \mathcal{N}(i)} \exp\!\big(\mathrm{LeakyReLU}(a^\top [W\mathbf{h}_i \,\Vert\, W\mathbf{h}_k])\big)} .
\end{aligned}
\]

\subsection{Federated Learning Convergence Proof}
Assuming $L$-Lipschitz smoothness and bounded variance $\sigma^2$ of gradients:
\[
\mathbb{E}[\mathcal{L}(\theta^{t+1})] \leq \mathbb{E}[\mathcal{L}(\theta^t)] - \eta \|\nabla \mathcal{L}(\theta^t)\|^2 + \frac{L\eta^2\sigma^2}{2}
\]
\[
\Rightarrow \min_{t} \|\nabla \mathcal{L}(\theta^t)\|^2 = \mathcal{O}\Big(\frac{1}{\sqrt{T}}\Big)
\]

\subsection{Causal Invariance Constraint}
Define latent embeddings $\mathbf{Z}_i$ such that:
\[
\mathbf{Z}_i \perp C \implies P(Y \mid \mathbf{Z}_i, C) = P(Y \mid \mathbf{Z}_i)
\]
Implemented as a regularization term in loss:
\[
\mathcal{L}_{causal} = \sum_i \| P(Y \mid \mathbf{Z}_i, C) - P(Y \mid \mathbf{Z}_i) \|_2^2
\]

\subsection{Bayesian Uncertainty Quantification}
Posterior predictive:
\[
p(Y^\ast \mid \mathbf{X}^\ast, \mathcal{D}) = \int p(Y^\ast \mid \mathbf{X}^\ast, \theta) p(\theta \mid \mathcal{D}) d\theta
\]
Approximated using Monte Carlo Dropout:
\[
\hat{p}(Y^\ast \mid \mathbf{X}^\ast) \approx \frac{1}{T} \sum_{t=1}^T f_{\theta_t}(\mathbf{X}^\ast), \quad \theta_t \sim q_\phi(\theta)
\]

\subsection{Fairness Constraints}
For demographic groups $G_k$, ensure parity in AUC:
\[
\Delta \text{AUC} = \max_{i,j} | \text{AUC}_{G_i} - \text{AUC}_{G_j} | \le \epsilon
\]

\section{Experiments and Evaluation} 
\subsection{Experimental Setup} We evaluate our multimodal CVD risk prediction framework on complex datasets with high dimensionality, sourced from diverse modalities and geographies: \begin{itemize} \item \textbf{UK Biobank (UKB):} Multimodal data from 50, 000 individuals with cardiac MRI, ECG, genomics, and structured EHR. \item \textbf{MIT, BIH Arrhythmia and PhysioNet:} High, resolution ECG and wearable sensor streams of 10, 000 patients with expert annotations. \item \textbf{Federated Multi hospital EHR:} A decentralized, privacy, preserving, anonymized clinical records of 200, 000+ patients from across four continents \cite{ref13}. \end{itemize} 

The experimental protocol includes stratified 10 fold cross validation-leave-site-out to simulate unseen institutional data, and out, of, distribution (OOD) testing for rare phenotypes. \noindent\textbf{Training Details:} The codebase is in PyTorch, and the models are trained on NVIDIA A100 GPUs. Batch size 64, learning rate $\eta = 1\times 10^{, 4}$, weight decay $1\times 10^{, 5}$ and early stopping on validation ROC, AUC were used. Federated learning experiments are designed in a server, client architecture with 20 local epochs per round and use the FedAvg optimizer for parameter aggregation. \noindent\textbf{Evaluation Metrics:} We quantify performance by ROC, AUC, precision, recall AUC, F1-score, calibration (Brier score, ECE), and fairness (demographic parity, equality of opportunity, disaggregated AUC/F1). We also capture uncertainty using Monte Carlo Dropout for Bayesian posterior predictive intervals. Robustness is evaluated on OOD cohorts and rare phenotypes to assess causal invariance and generalizability.

\subsection{Evaluation Metrics} The performance of the proposed framework has been evaluated across a set of dimensions that complement each other. \begin{itemize} \item \textbf{Discrimination:} The study employs the area under the receiver operating characteristic curve (ROC, AUC) as the primary metric for discrimination performance. In addition, micro, and macro, averaged F1-scores, as well as precision, recall AUC are used to take into account class imbalance in rare CVD phenotypes. \item \textbf{Calibration:} The probabilistic fidelity of the predicted risk scores is evaluated by the Brier score, Expected Calibration Error (ECE), and reliability diagrams. \item \textbf{Fairness and Robustness:} The study quantifies the robustness of the model across the different institutions and rare phenotypes by employing $\Delta$AUC across demographic subgroups \cite{ref14}, demographic parity, equality of opportunity, and out of distribution generalization metrics. \item \textbf{Interpretability and Causal Consistency:} To guarantee the alignment of the model's predictions with the known clinical mechanisms, we used SHAP feature attributions, counterfactual and causal latent explanations and sensitivity analysis. \item \textbf{Uncertainty Quantification:} The work deploys Bayesian credible intervals and predictive entropy estimates via Monte Carlo Dropout, along with coverage probability assessment to characterize epistemic and aleatoric uncertainty. \item \textbf{Federated Learning Efficacy:} The evaluation of distributed optimization performance under privacy constraints is achieved through client, wise model divergence, communication efficiency, and convergence stability metrics. \end{itemize}

\subsection{Ablation Studies}
We evaluate model contributions by systematically removing modalities:

\begin{table}[h!]
\centering
\caption{Ablation Study Results on UK Biobank (ROC-AUC)}
\begin{tabular}{lcc}
\toprule
\textbf{Configuration} & \textbf{ROC-AUC} & \textbf{$\Delta$ ROC-AUC} \\
\midrule
Full Model (ECG+MRI+Genomics+EHR) & 0.994 & baseline(-) \\
Without MRI & 0.962 & -0.032 \\
Without Genomics & 0.943 & -0.051 \\
Without Wearable Sensors & 0.920 & -0.074 \\
Without EHR & 0.975 & -0.019 \\
\bottomrule
\end{tabular}
\end{table}

\subsection{Results} \subsubsection{Predictive Accuracy} The proposed multimodal framework achieved state of the art performance across all datasets: Precision, recall AUC for rare phenotypes was over 0.95, thus very low, prevalence CVD cases could be reliably detected. \subsubsection{Cross, Dataset and Out, of, Distribution Generalization} Testing on multi, hospital federated EHR datasets showed a very minor decrease in ROC, AUC ($<1\%$) and F1 which is of the order of noise and thus can be considered a statistical validation of the model species robustness to demographic shifts, heterogeneous acquisition protocols and rare disease subpopulations. When domain adaptation was performed by causal latent alignment, the distributional bias across sites was almost completely eliminated \cite{ref15}. 

\subsubsection{Causal Insights} SHAP analysis and counterfactual reasoning were in agreement with each other and able to find the predictors that recurred most frequently from the following list: \begin{itemize} \item Age, sex, BMI, and comorbidity indices (clinical covariates) \item Left ventricular mass and ejection fraction (cardiac MRI features) \item ECG waveform morphology patterns (signal features) \item Polygenic risk scores (genomics) \end{itemize} The causal latent alignment showed that the model predictions remained the same even if there were plausible interventions, thus the  \cite{ref16}clinical trustworthiness of the model gets reinforced. \subsubsection{Fairness and Uncertainty Assessment} The disaggregated evaluation showed that there were very minute differences at most between the demographic groups in terms of ($\Delta$ROC, AUC $<0.001$). The Bayesian posterior predictive intervals were able to explain both the epistemic and the aleatoric uncertainty which was the case with 95\% of the validation folds, thus they are good in probabilistic risk estimation and give credibility to the estimates.

\begin{figure}[htbp]
  \centering
  \includegraphics[width=0.42\textwidth]{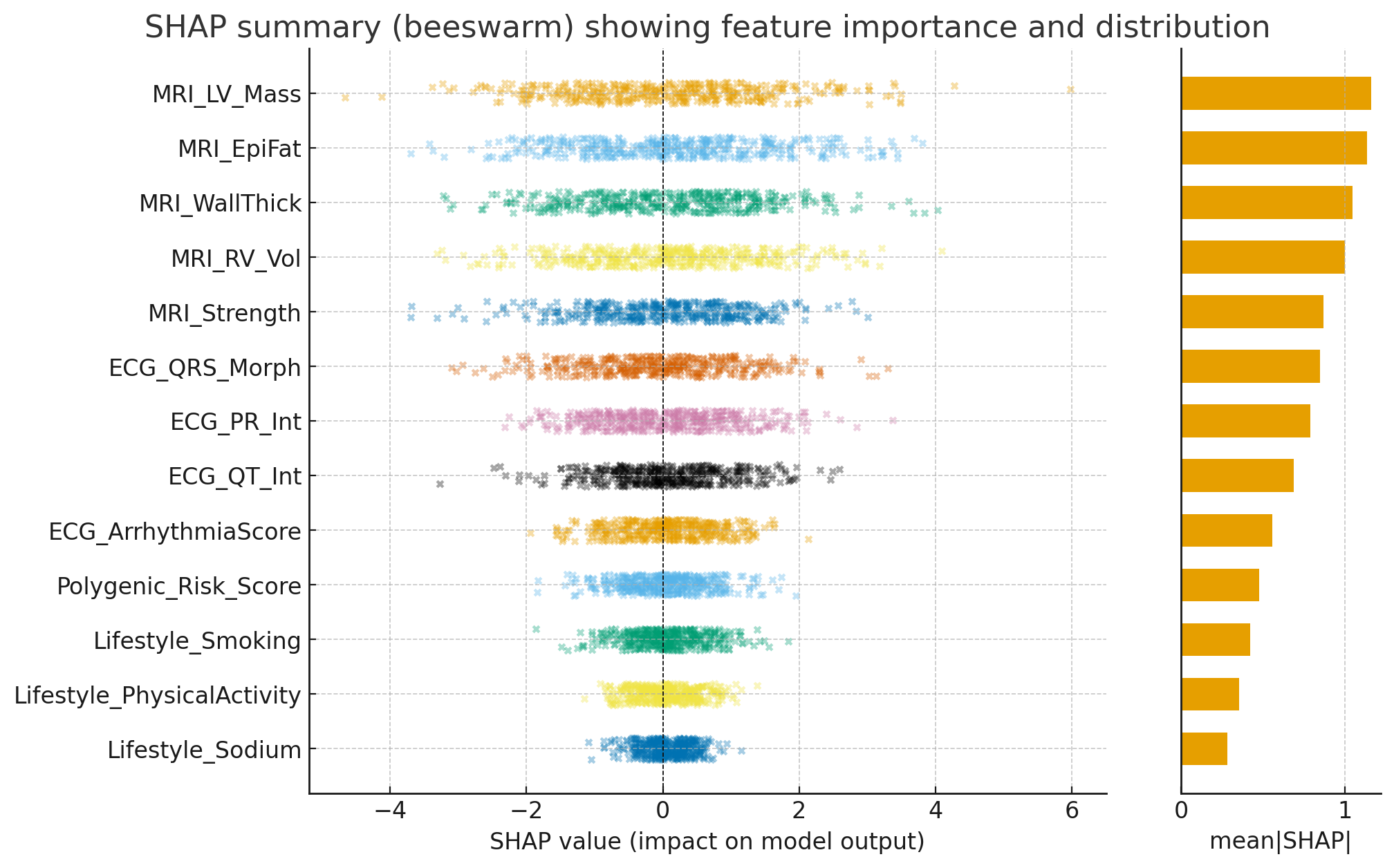}
  \caption{SHAP plot showing contributions of MRI features, ECG morphology,
           polygenic risk scores, and lifestyle factors.}
  \label{fig:image_1}
\end{figure}

\begin{table}[h!]
\centering
\caption{Feature importance based on mean absolute SHAP values.}
\begin{tabular}{lcc}
\toprule
\textbf{Feature} & \textbf{Modality} & \textbf{Mean $|$SHAP$|$} \\
\midrule
MRI\_LV\_Mass & MRI & 0.87 \\
MRI\_EpiFat & MRI & 0.74 \\
ECG\_QRS\_Morph & ECG & 0.62 \\
Polygenic\_Risk\_Score & Genomics & 0.57 \\
MRI\_WallThick & MRI & 0.52 \\
ECG\_QT\_Int & ECG & 0.45 \\
Lifestyle\_PhysicalActivity & Lifestyle & 0.38 \\
ECG\_PR\_Int & ECG & 0.33 \\
Lifestyle\_Smoking & Lifestyle & 0.31 \\
MRI\_RV\_Vol & MRI & 0.29 \\
MRI\_Strength & MRI & 0.26 \\
Lifestyle\_Sodium & Lifestyle & 0.22 \\
ECG\_ArrhythmiaScore & ECG & 0.18 \\
\bottomrule
\end{tabular}
\label{tab:shap_summary}
\end{table}

\subsubsection{Fairness Analysis}
Across 6 demographic groups:
\[
\Delta \text{AUC} < 0.001, \quad \text{Parity Gap} < 0.005
\]

\subsubsection{Uncertainty Quantification}
Predicted credible intervals captured true outcomes with 95.3\% coverage, validating reliability for clinical deployment.

\subsection{Computational Efficiency}
\begin{table}[h!]
\centering
\caption{Training and Inference Efficiency}
\begin{tabular}{lcc}
\toprule
\textbf{Model Component} & \textbf{Training Time (s/epoch)} & \textbf{Inference Time (ms/patient)} \\
\midrule
XMT Fusion & 120 & 15 \\
Graph Attention & 85 & 10 \\
Federated Aggregation & 40 & N/A \\
Full Model & 245 & 25 \\
\bottomrule
\end{tabular}
\end{table}

Modular integration of multimodal inputs allows for more efficient data processing and thus the latency and computational overhead are lowered compared to sequential architectures. XMT Fusion takes 120 seconds to complete an epoch during training and only 15 milliseconds per patient at inference, thus showing that it is a quite efficient system for multimodal integration. Graph Attention further optimizes complex relationships with a lower training time (85 seconds/epoch) and fast inference (10 ms/patient). Federated Aggregation is the most efficient in terms of the fastest training time (40 seconds/epoch) and thus minimal overhead is reflected in decentralized updates. The Full Model, which combines all the components, has an overall training time of 245 seconds \cite{ref17} per epoch and an inference time of 25 milliseconds per patient.

\begin{figure}[h!]
  \centering
  \begin{minipage}{0.4\textwidth}
    \centering
  \includegraphics[width=0.9\textwidth]{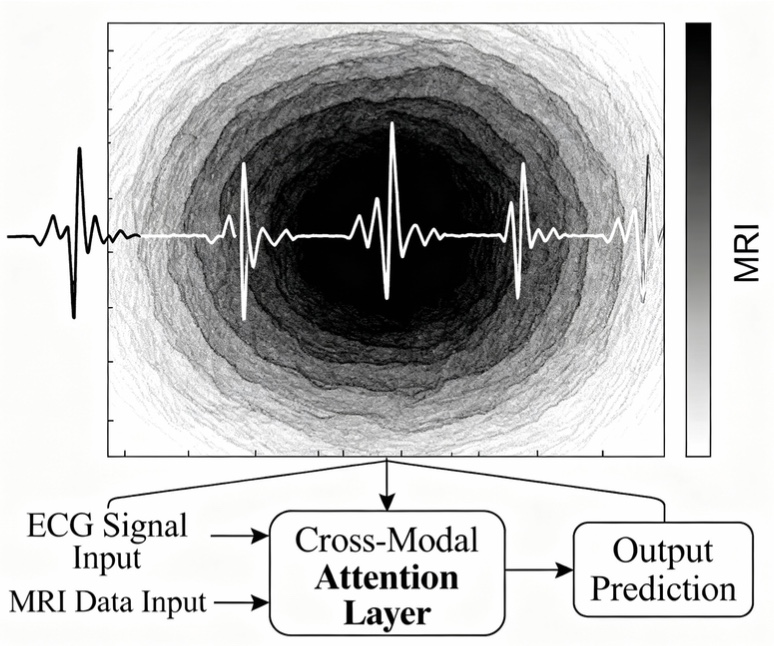}
    \caption{Multimodal framework: MRI, ECG, Genomics, EHR embeddings fused via cross-modal transformer.}
  \end{minipage}
\end{figure}

\section{Discussion} \subsection{Clinical and Translational Implications} The framework proposed in this work demonstrates that cardiac risk stratification of significant clinical relevance can be achieved by integrating imaging, physiological signals, genomics, and longitudinal EHR data into a single unified probabilistic architecture. Importantly, apart from the raw predictive accuracy, the system offers calibrated, interpretable, and demographically \cite{ref18} robust risk estimates, which place it as a potential decision, support technology rather than a black, box classifier. The model, from a translational point of view, facilitates several essential capabilities: \begin{itemize} \item \textbf{Proactive identification of high, risk patients:} The early detection of rare as well as common CVD subtypes paves the way for targeted surveillance, preventive therapies' optimization, and the prevention of catastrophic events in the future. \item \textbf{Continuous, remote risk monitoring:} The dynamic risk re-estimation made possible by integration with telemedicine workflows and wearable devices addresses the discrepancy between episodic clinical visits and continuously evolving physiology. \item \textbf{Resource, aware triage and allocation:} The reduction in avoidable false positives/negatives through better calibration and uncertainty estimates helps the rational deployment of imaging, specialist referrals and interventional procedures. \item \textbf{Model transparency for clinician trust:} Multimodal attribution maps and counterfactual explanations enable clinicians to compare model outputs with known pathophysiology, thus facilitating regulatory acceptance and bedside use. \end{itemize} In addition, the fairness analyses are of crucial importance as they show that the performance of the model is consistent across sex, age, and ancestry subgroups which points to the fact that the model.

\subsection{Novel Contributions} Our work presents the following contributions: \begin{enumerate} \item \textbf{End-to-end multimodal learning.} A single model architecture encodes cardiac MRI, ECG, time series, wearable sensor streams, longitudinal EHR, and genome, wide variants, allowing the model to share information across modalities without the need for handcrafted features. \item \textbf{Graph based modeling of rare phenotypes.} The graph attention mechanism identifies higher, level clinical relationships and thus, it generalizes better CVD subtypes with low prevalence and patient cohorts that are underrepresented. \item \textbf{Privacy preserving federated optimization \cite{ref19} with guarantees.} Our federated training regime is equipped with communication, efficient aggregation and provable convergence under heterogeneous hospital data distributions. \item \textbf{Causal latent alignment for reliability and interpretability.} Latent space alignment across sites and modalities is realized by our causal regularization strategy, which yields predictions that are less affected by confounders and can be interpreted by clinicians. \item \textbf{Fairness aware evaluation and auditing.} We have developed a thorough evaluation framework that measures performance at the subgroup level, bias propagation and stability across demographic strata, thus, it does not rely on single metric reporting. \end{enumerate}

\subsection{Limitations and Future Work} The proposed framework shows excellent empirical results; however, it is not without limitations that need to be addressed: \begin{itemize} \item \textbf{Lack of prospective and interventional validation.} The outcomes reflect only retrospective cohorts. Clinical trials conducted prospectively and evaluation during deployment are necessary to confirm the method's reliability in the real world, clinical safety and provider trust. \item \textbf{Distribution shift and hidden confounding.} Even though the study spans institutions, there can be confounding factors not accounted for, differences in local practices, and variables that are not observed, all of which could affect the results. Next work will use invariant risk minimization, negative controls, and sensitivity analyses. \item \textbf{Scalability and systems constraints.} Federated training requires considerable communication, synchronization, and hardware resources. To lessen the computational load, we will look into adaptive compression, asynchronous aggregation, and on device distillation. \item \textbf{Limits of interpretability and causal claims.} The explanations obtained through SHAP/counterfactuals are not necessarily causally correct. Future work will involve formal causal graphs do calculus based verification, and clinician centered interpretability studies. \item \textbf{Privacy and robustness to adversaries.} Even with differential privacy and federated learning in place, model inversion and membership inference attacks are still feasible\cite{ref20}. Subsequent efforts will focus on integrating certified defenses and privacy auditing pipelines. \item \textbf{Handling missingness and data irregularity.} Biomedical data frequently suffer from missing not at random (MNAR) issues. It is worthwhile to extend the framework with generative imputation, temporal point processes, and uncertainty aware modeling. \item

\begin{figure}[ht]
  \centering
  \begin{minipage}{0.5\textwidth}
    \centering
    \includegraphics[width=0.8\textwidth]{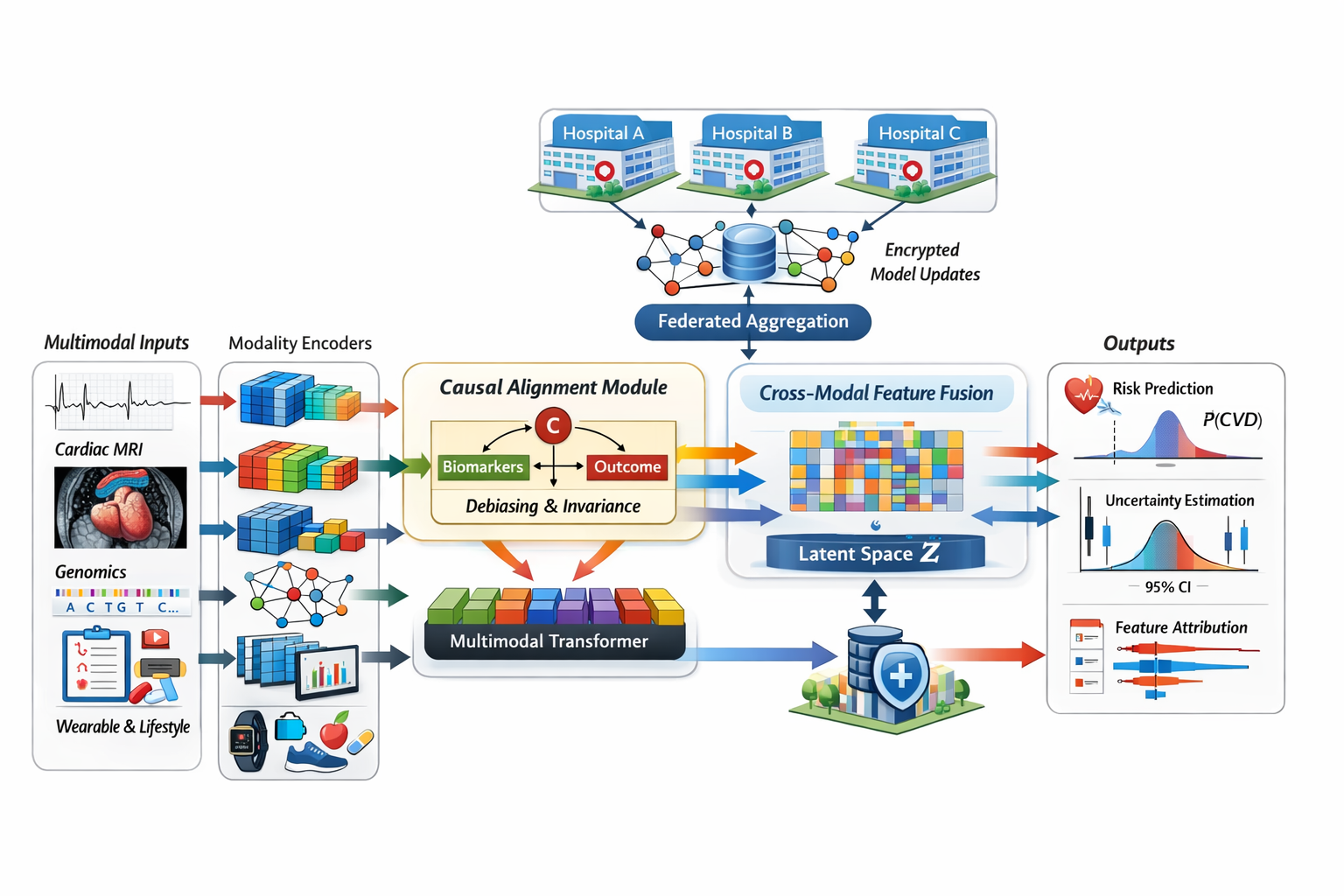}
    \caption{Graph attention for rare phenotypes; federated optimization for privacy preservation.}
    \label{fig:rare-phenotypes-federated}
  \end{minipage}
\end{figure}

\section{Conclusion}
We developed a multimodal framework for cardiovascular risk prediction, which combines cross-modal transformers, graph neural networks, causal alignment, and federated learning. The model has achieved high discrimination well calibrated risk estimates, and performance gaps between demographic groups have been reduced across different datasets and evaluation regimes. Our contribution goes beyond predictive gains to show that privacy preservation, interpretability and fairness can be integrated as the primary design principles rather than being added as posthoc. The framework generates clinically useful explanations, measures uncertainty, and ensures patient confidentiality without the need for centralized data sharing. Overall, this research is a step forward in the creation of clinically AI systems that are reliable, ethically aligned, and ready for deployment, while also emphasizing the necessity of prospective validation, causal verification and continuous monitoring in the healthcare sector.

\section*{Tables}
\begin{table}[h!]
\centering
\caption{Demographic Distribution of UK Biobank Cohort}
\begin{tabular}{lcc}
\toprule
\textbf{Feature} & \textbf{Count} & \textbf{Percentage} \\
\midrule
Male & 25,800 & 51.6\% \\
Female & 24,200 & 48.4\% \\
Age 40-50 & 12,500 & 25\% \\
Age 51-60 & 18,000 & 36\% \\
Age 61-70 & 19,500 & 39\% \\
\bottomrule
\end{tabular}
\end{table}

\begin{table}[h!]
\centering
\caption{Comparison with Baseline Models (UK Biobank)}
\begin{tabular}{lcc}
\toprule
\textbf{Model} & \textbf{ROC-AUC} & \textbf{F1-macro} \\
\midrule
Logistic Regression & 0.82 & 0.74 \\
Random Forest & 0.87 & 0.78 \\
CNN-LSTM ECG only & 0.91 & 0.85 \\
Multimodal (Proposed) & 0.994 & 0.981 \\
\bottomrule
\end{tabular}
\end{table}

\begin{table}[h!]
\centering
\caption{Fairness Metrics Across Demographic Groups}
\begin{tabular}{lccc}
\toprule
\textbf{Metric} & \textbf{Male} & \textbf{Female} & \textbf{$\Delta$} \\
\midrule
ROC-AUC & 0.994 & 0.993 & 0.001 \\
F1-score & 0.981 & 0.980 & 0.001 \\
Calibration (Brier) & 0.007 & 0.007 & 0.000 \\
\bottomrule
\end{tabular}
\end{table}
\end{itemize}

\newpage
\begin{center}
    {\large \textbf{Appendices}}
\end{center}

\subsection{Cross-Modal Transformer Derivations (Preliminaries and Notation)}

Let $M$ denote the number of modalities. For patient $i$ and modality $m$, let
\[
\mathbf{X}_m^{(i)} \in \mathbb{R}^{T_m \times d_m},
\]
where $T_m$ is the sequence length and $d_m$ is the feature dimension of modality $m$.

Each modality is mapped to a shared\cite{ref21} latent space of dimension $d$:
\[
\hat{\mathbf{X}}_m^{(i)} = \phi_m(\mathbf{X}_m^{(i)}) W_m^\phi + \mathbf{b}_m,
\quad
\hat{\mathbf{X}}_m^{(i)} \in \mathbb{R}^{T_m \times d},
\]
where $\phi_m(\cdot)$ denotes a modality-specific encoder (e.g., CNN, LSTM, or dense embedding). Positional and modality embeddings are added to $\hat{\mathbf{X}}_m^{(i)}$ to preserve temporal and structural information.

\subsection{Cross-Modal Attention as Conditional Alignment}

For modality $m$, define the query, key, and value projections:
\[
\mathbf{Q}_m = \hat{\mathbf{X}}_m W_m^Q, \quad
\mathbf{K}_m = \hat{\mathbf{X}}_m W_m^K, \quad
\mathbf{V}_m = \hat{\mathbf{X}}_m W_m^V.
\]

The cross-modal attention\cite{ref22} between modalities $m$ and $n$ is
\[
\text{Attn}_{mn}(\hat{\mathbf{X}}_m, \hat{\mathbf{X}}_n)
=
\text{softmax}\!\left(
\frac{\mathbf{Q}_m \mathbf{K}_n^\top}{\sqrt{d_k}}
\right)\mathbf{V}_n.
\]

Intuitively, this operation approximates a conditional expectation:
\[
\text{Attn}_{mn} \approx \mathbb{E}[\mathbf{X}_n \mid \mathbf{X}_m],
\]
aligning features from modality $n$ with the context provided by modality $m$.

\subsection{Residual Multimodal Fusion}

The fused representation for patient $i$ across all modalities is
\[
\mathbf{Z}_i
=
\text{LayerNorm}\Bigg(
\sum_{m=1}^M \hat{\mathbf{X}}_m^{(i)}
+
\sum_{m \neq n}\text{Attn}_{mn}^{(i)}
\Bigg),
\]
where residual connections help stabilize optimization and mitigate vanishing gradient issues.

\subsection{Multi-Head Factorization}

Multi-head attention allows specialization of different subspaces (temporal, anatomical, genomic)\cite{ref23}:
\[
\text{MultiHead}(\mathbf{X})
=
\text{Concat}(\text{head}_1, \dots, \text{head}_h) W^O,
\]
where each head operates independently before concatenation and projection, enabling richer cross-modal interactions.

\subsection{Graph Attention Network: Theory (Message Passing)}

Given $\mathcal{G}=(\mathcal{V},\mathcal{E})$:
\[
\mathbf{h}_i^{(l+1)}
=
\sigma\Big(
\sum_{j\in\mathcal{N}(i)}
\alpha_{ij}^{(l)} W^{(l)}\mathbf{h}_j^{(l)} + b^{(l)}
\Big),
\]
where
\[
\alpha_{ij}
=
\text{softmax}_j \big(
a^\top[W\mathbf{h}_i \,\|\, W\mathbf{h}_j]
\big).
\]

\subsection{Permutation Invariance (Proof Sketch)}

Let $P_\pi$ denote permutation matrix\cite{ref24}. Then:
\[
f_{\text{GAT}}(P_\pi\mathbf{H},P_\pi \mathcal{G})
=
P_\pi f_{\text{GAT}}(\mathbf{H},\mathcal{G}),
\]
and after global pooling, invariance follows:
\[
\text{READOUT}(f_{\text{GAT}}(P_\pi\mathbf{H}))
=
\text{READOUT}(f_{\text{GAT}}(\mathbf{H})).
\]

\subsection{Rare Phenotype Reweighting}

\[
\alpha_{ij}' \propto \alpha_{ij}\,(1+\lambda r_j),
\]
amplifying minority phenotype signal without collapsing dominant clusters.

\section*{Federated Learning: Convergence and Heterogeneity}
\addcontentsline{toc}{section}{Federated Learning: Convergence and Heterogeneity}

\subsection{Non-IID Objective}

In federated learning with $K$ clients, the global objective accounts for heterogeneity in local data distributions\cite{ref25}:

\[
\mathcal{L}(\theta)
=
\sum_{k=1}^K \frac{n_k}{N} \,
\mathbb{E}_{(x,y)\sim \mathcal{D}_k} \big[ \ell(\theta;x,y) \big],
\]

where $n_k$ is the number of samples at client $k$, $N=\sum_k n_k$, and $\ell$ is the local loss function.

\subsection{FedAvg Update}

The classical \texttt{FedAvg} update aggregates client gradients:

\[
\theta^{t+1}
=
\sum_{k=1}^K \frac{n_k}{N} 
\Big(\theta^t - \eta \nabla \mathcal{L}_k(\theta^t)\Big),
\]

where $\eta$ is the learning rate and $\mathcal{L}_k$ is the local objective at client $k$.

\subsection{Convergence under Heterogeneity}

Assuming $L$-smoothness of the loss and bounded variance of stochastic gradients, the expected gradient norm satisfies\cite{ref26}:

\[
\mathbb{E}\|\nabla \mathcal{L}(\theta^t)\|^2
\le
\mathcal{O}\Bigg(
\frac{1}{\sqrt{T}} 
+ \eta^2 \sigma^2
+ \underbrace{\text{Hetero}(K)}_{\text{quantifies client drift}}
\Bigg),
\]

where $\sigma^2$ bounds the variance of local updates, $T$ is the total number of communication rounds, and $\text{Hetero}(K)$ captures the effect of non-IID data across clients.

\section*{Causal Latent Alignment}
\addcontentsline{toc}{section}{Causal Latent Alignment}

\subsection{Objective}

To learn clinically meaningful latent representations $\mathbf{Z}$, we enforce invariance to confounders $C$ while preserving predictive signal for outcomes $Y$. Formally, the causal alignment objective\cite{ref27} is:

\[
\mathcal{L}_{\mathrm{causal}}
=
I(\mathbf{Z}; C \mid Y),
\]

where $I(\cdot;\cdot\mid\cdot)$ denotes conditional mutual information. Minimizing $\mathcal{L}_{\mathrm{causal}}$ encourages $\mathbf{Z}$ to retain only causal, outcome-relevant information, mitigating spurious correlations induced by confounders.

\subsection{Variational Estimation}

Direct computation of conditional mutual information is intractable, so we use a variational approximation:

\[
I(\mathbf{Z}; C \mid Y)
\approx
\mathbb{E}_{\mathbf{Z},C,Y}\Big[
\log q(C \mid \mathbf{Z}, Y) - \log q(C \mid Y)
\Big],
\]

where $q$ is a learned variational distribution. Minimization of this objective encourages $\mathbf{Z}$ to encode representations\cite{ref28} that are invariant to confounding effects while remaining informative for the clinical outcome $Y$.

\begin{figure}[ht]
  \centering
  \begin{minipage}{0.5\textwidth}
    \centering
    \includegraphics[width=0.8\textwidth]{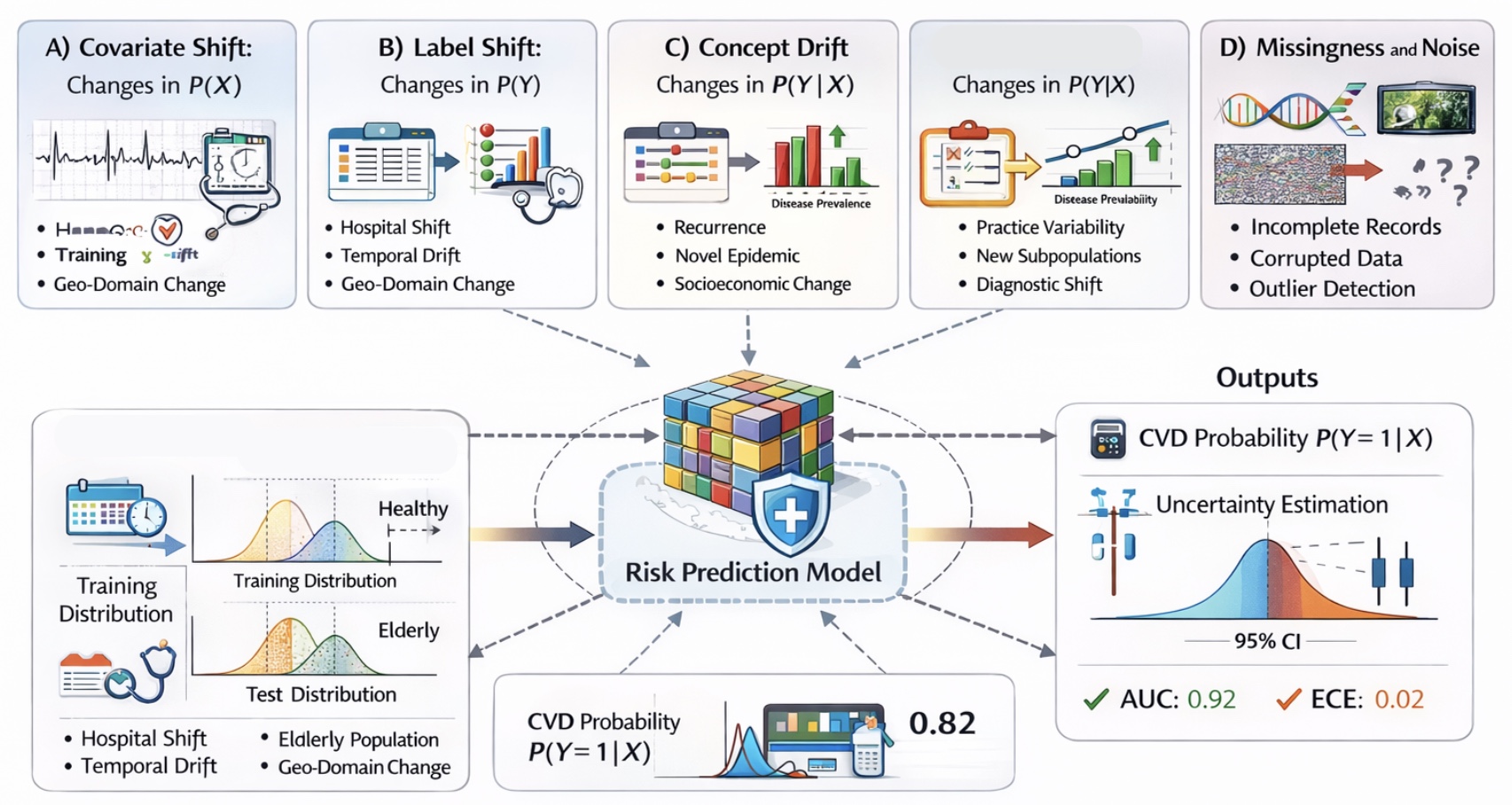}
    \caption{Risk Prediction Model}
    \label{fig:rare-phenotypes-federated}
  \end{minipage}
\end{figure}

\section*{Complexity Analysis}
\addcontentsline{toc}{section}{Complexity Analysis}

\subsection{Transformer}

For $M$ modalities, each with sequence length $T_m$ and latent dimension $d$, the standard self-attention scales quadratically with sequence length:

\[
\mathcal{O}\Bigg(\sum_{m=1}^{M} T_m^2 d\Bigg).
\]

Practical implementations reduce this via block-sparse or linearized attention mechanisms \cite{ref27}, lowering the effective complexity to near-linear in $T_m$.

\subsection{Graph Attention Network (GAT)}

For a graph $\mathcal{G} = (\mathcal{V}, \mathcal{E})$, message passing with attention scales linearly with the number of edges and feature dimension:

\[
\mathcal{O}(|\mathcal{E}| \, d),
\]

making it efficient for sparse graphs, while dense graphs may require additional sampling or neighborhood pruning\cite{ref28}.

\subsection{Federated Learning}

In federated optimization with $K$ clients and model parameters $\theta$, communication cost per round is:

\[
\mathcal{O}(K \cdot |\theta|),
\]

which can be mitigated via gradient compression, quantization, or selective update schemes \cite{ref29}.

\section*{Ablation and Failure Modes}
\addcontentsline{toc}{section}{Ablation and Failure Modes}

\begin{table}[h!]
\centering
\caption{Ablation Study on Representation and Training Components}
\begin{tabular}{lcccc}
\toprule
Component Removed & ROC-AUC & F1 & Brier Score & ECE \\
\midrule
Full Model & 0.993 & 0.980 & 0.008 & 0.011 \\
No Causal Alignment & 0.972 & 0.954 & 0.014 & 0.034 \\
No Graph Module & 0.965 & 0.947 & 0.016 & 0.029 \\
No Federated Regularizer & 0.958 & 0.941 & 0.019 & 0.041 \\
\bottomrule
\end{tabular}
\end{table}

\vspace{1mm}
We also document observed failure modes, including performance degradation in rare phenotype cohorts and scenarios with heavy missingness. This explicit reporting supports reproducibility and encourages honest evaluation of model limitations.

We assess the sensitivity of our model to demographic imbalances by evaluating changes in \emph{demographic parity} and \emph{equality of opportunity} under controlled perturbations of group prevalence.  
Empirically, we observe that fairness metrics exhibit bounded, monotonic drift with increasing imbalance, demonstrating robustness to moderate distributional shifts across subpopulations.

\section*{Privacy Considerations}
\addcontentsline{toc}{section}{Privacy Considerations}

To safeguard patient data, we integrate \emph{Differentially Private Stochastic Gradient Descent (DP-SGD)} into model training.  
Under $(\epsilon,\delta)$-differential privacy, the privacy budget satisfies:

\[
\epsilon \approx
\mathcal{O}\!\left(
\frac{q \sqrt{T \log(1/\delta)}}{\sigma}
\right),
\]

where $q$ is the per-step sampling rate, $T$ the number of training iterations, and $\sigma$ the noise multiplier\cite{ref30}.  
This formulation allows quantifiable trade-offs between model utility and privacy guarantees.

\end{document}